\documentclass{article}
\usepackage{spconf}
\usepackage{amsmath}
\usepackage{graphicx}
\usepackage{hyperref}

\title{Modularity-Free Conflict-Averse Training for Generalized PINNs}
\name{
Heejo Kong\textsuperscript{\rm 1},
Beomchul Park\textsuperscript{\rm 2},
Sung-Jin Kim\textsuperscript{\rm 2},
Seong-Whan Lee\textsuperscript{\rm 2}
\thanks{
This work was supported by Institute for Information \& Communications Technology Planning \& Evaluation (IITP) grant funded by the Korea government (MSIT) (No. RS-2019-II190079), by the Korea Institute for Advancement of Technology (KIAT) grant funded by the Ministry of Trade, Industry and Energy (MOTIE) (No. RS-2025-12572968).
}
}
\address{
\textsuperscript{\rm 1}Department of Brain and Cognitive Engineering, Korea University, Seoul, Republic of Korea \\
\textsuperscript{\rm 2}Department of Artificial Intelligence, Korea University, Seoul, Republic of Korea
\\
\{hj\_kong, sea\_breeze88, s\_j\_kim, sw.lee\}@korea.ac.kr
}
\usepackage{booktabs}
\usepackage{multirow}
\usepackage{amsmath}
\usepackage{amsfonts}
\usepackage{bbm}
\usepackage[table]{xcolor}
\usepackage{algpseudocode}
\usepackage{etoolbox}
\makeatletter
\newcommand*{\algrule}[1][\algorithmicindent]{%
  \makebox[#1][l]{%
    \hspace*{.2em}%
    \vrule height .90\baselineskip depth .25\baselineskip
  }
}
\newcount\ALG@printindent@tempcnta
\def\ALG@printindent{%
  \ifnum \theALG@nested>0%
    \ifx\ALG@text\ALG@x@notext%
    \else
      \unskip
      \addvspace{-1pt}%
      \ALG@printindent@tempcnta=1
      \loop
        \algrule[\csname ALG@ind@\the\ALG@printindent@tempcnta\endcsname]%
        \advance \ALG@printindent@tempcnta 1
      \ifnum \ALG@printindent@tempcnta<\numexpr\theALG@nested+1\relax
      \repeat
    \fi
  \fi
}
\patchcmd{\ALG@doentity}{\noindent\hskip\ALG@tlm}{\ALG@printindent}{}{\errmessage{failed to patch}}
\makeatother

\begin{document}
\ninept
\maketitle
\begin{abstract}
Physics-informed neural networks (PINNs) have become a powerful framework for solving PDEs by embedding physical laws into differentiable objectives.
Despite their advances, training PINNs remains fragile: recent conflict-averse optimization schemes alleviate gradient interference between residual and boundary losses, but we show that their effectiveness deteriorates as model capacity increases.
In this paper, we identify a capacity-induced failure mode, where overparameterized networks undergo functional modularity, self-partitioning into task-exclusive modules that suppress cross-objective interaction and hinder convergence toward Pareto-stationary points.
To address this issue, we propose a novel framework, Modular-Sparsity Synchronization (ModSync), which integrates structural optimization into conflict-averse training by penalizing task-exclusive connections while preserving interaction-promoting pathways.
Extensive experiments across diverse PDE benchmarks demonstrate that ModSync consistently prevents capacity-driven failures, sustains robust cross-objective coupling, and achieves state-of-the-art accuracy. Codes are available at \url{https://github.com/heejokong/ModSync}.
%
\end{abstract}
\begin{keywords}
Physics-informed neural networks, multi-task learning, conflict-averse optimization, dynamic sparse training
\end{keywords}
\section{Introduction}
Physics-informed neural networks (PINNs) \cite{pinn_1, pinn_2} have emerged as a powerful approach for solving partial differential equations (PDEs) that underlie complex phenomena across science and engineering \cite{pde_1, pde_2, la2former}.
Compared with traditional numerical solvers \cite{fem, fdm}, PINNs embed physical laws directly into the training objective, using neural networks as differentiable ansatzes to approximate the underlying physics.
This paradigm enables mesh-free training and scalable approximation, making PINNs particularly effective for high-dimensional systems where traditional methods struggle.

Despite their promise, training PINNs remains challenging due to the unclear nature of training pathologies \cite{failure_1, failure_2, failure_3, failure_4, failure_5}.
These challenges often lead to failures to learn correct solutions, even in relatively simple settings, thereby hindering their broader applicability.
Recent works \cite{dcgd, config}, inspired by multi-task learning (MTL) \cite{mgda, pcgrad}, have proposed conflict-averse training to mitigate negative interference between the PDE residual and data loss via gradient manipulation, yielding notable gains across diverse physics scenarios.
However, these studies largely focus on gradient refinement while leaving other crucial aspects of PINN optimization underexplored.

Building on the above limitations of gradient-centric remedies, we observe that as model capacity increases, existing conflict-averse methods can enter systematic failure modes.
As shown in Fig. 1, beyond a capacity threshold, some even underperform vanilla PINNs trained with Adam \cite{adam}, \textit{i.e.}, training that does not explicitly account for conflict between the objectives.
Through the lens of functional modularity \cite{fm_1, fm_2, fm_3}, we interpret this phenomenon as a capacity-induced shortcut whereby a single overparameterized network self-partitions into task-exclusive modules, allowing the PDE residual and data objectives to be optimized in isolation.
Such modular segregation suppresses cross-objective interaction and impedes convergence to Pareto-stationary points \cite{mgda, pareto}, thereby undermining the intended benefits of conflict-averse training.

\begin{figure}[t]
    \centering
\includegraphics[width=0.90\linewidth]{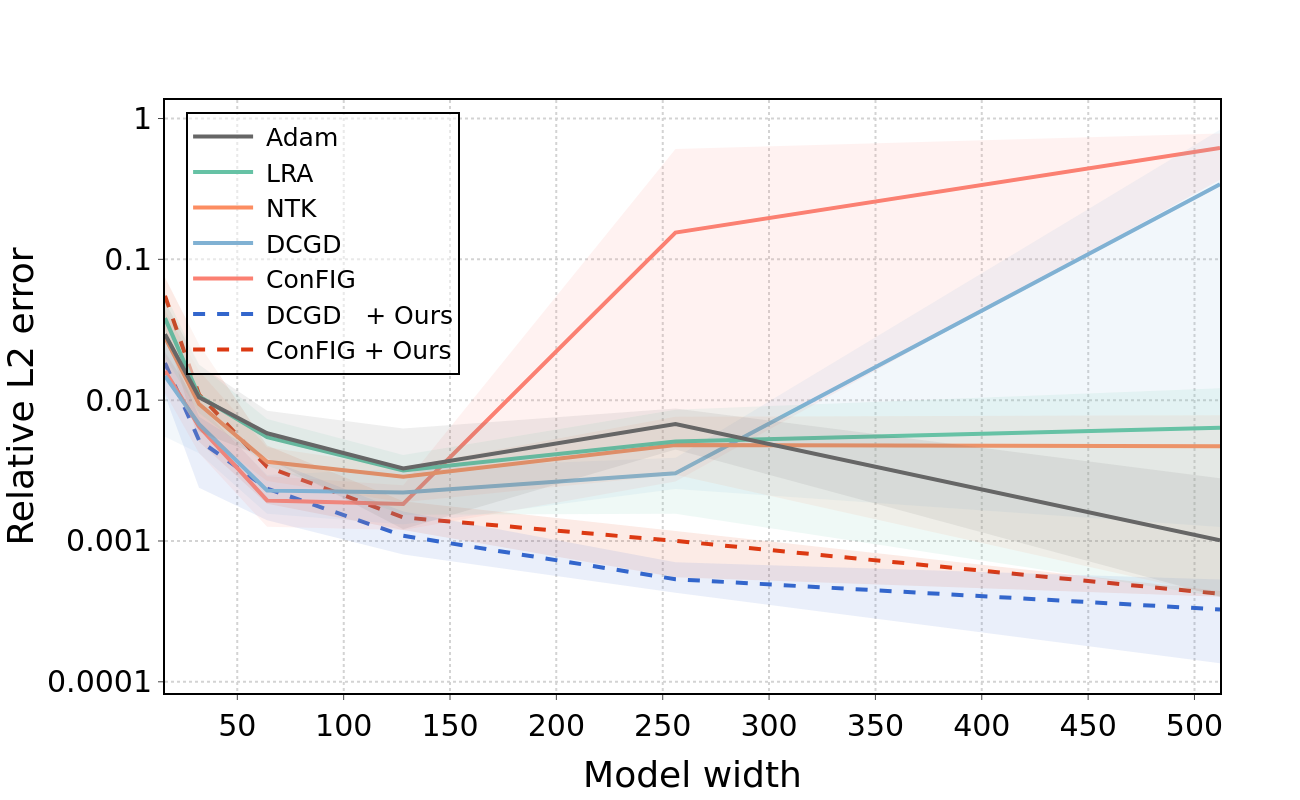}
\vspace{-0.2cm}
\caption{Effect of model capacity on PINN performance for the Burgers’ equation. With increasing width, existing conflict-averse methods (DCGD and ConFIG) suffer from systematic failures and can underperform vanilla PINNs (ADAM), while the proposed method alleviates these degradations.}
\vspace{-0.5cm}
\end{figure}
%
To address this drawback, we propose a novel framework, Modular-Sparsity Synchronization (ModSync), that incorporates structural optimization into conflict-averse training.
Our key idea is to identify network connections that become task-exclusive with respect to PINN objectives and penalize them, while preserving interaction-promoting pathways.
Specifically, our method dynamically optimizes both network parameters and structure by pruning regions where the two objectives evolve independently while preserving connections that foster interaction.
This approach mitigates unintended effects of functional modularity, sustains robust cross-objective coupling, and promotes convergence toward Pareto-stationary points.
Across diverse PDE benchmarks, we demonstrate that the proposed ModSync prevents—or substantially reduces—the failure modes exhibited by existing conflict-averse methods.
\section{Background}
\subsection{Preliminaries}
\noindent\textbf{Physics-informed neural networks} \cite{pinn_1}
are designed to solve partial differential equations (PDEs) by incorporating physical constraints directly into the training process.
Let $\Omega \in \mathbb{R}^{d}$ represent the domain, and $\partial\Omega$ is its boundary.
We consider nonlinear PDEs of the form:
\begin{equation}
\begin{aligned}
& \mathcal{N}[u(x), x]=f(x), \quad x \in \Omega, \\
& \mathcal{B}[u(x), x]=g(x), \quad x \in \partial \Omega,
\end{aligned}
\end{equation}
where $\mathcal{N}$ and $\mathcal{B}$ denote a nonlinear differential operator and a boundary condition operator, respectively.
PINNs approximate the solution $u(x)$ using a neural network $u(x;\theta)$ parameterized by $\theta$, trained to minimize two losses, the residual loss $\mathcal{L}_{r}(\theta)$ and boundary loss $\mathcal{L}_{b}(\theta)$, defined as:
\begin{equation}
\begin{aligned}
& \mathcal{L}_{r}(\theta)=\frac{1}{|N_{r}|} \sum_{x_{r}^{i} \in N_{r}} \left|\mathcal{N}[u(x_{r}^{i}; \theta), x_{r}^{i}]-f(x_{r}^{i})\right|^{2}, \\
& \mathcal{L}_{b}(\theta)=\frac{1}{|N_{b}|} \sum_{x_{b}^{i} \in N_{b}} \left|\mathcal{B}[u(x_{b}^{i}; \theta), x_{b}^{i}]-g(x_{b}^{i})\right|^{2}, 
\end{aligned}
\end{equation}
where $N_{r}$ and $N_{b}$ denote the sets of sampling points within $\Omega$ and on $\partial\Omega$, respectively.
$\mathcal{L}_{r}(\theta)$ quantifies the extent to which the predicted solution violates the governing PDE, while $\mathcal{L}_{b}(\theta)$ measures the degree of violation of the boundary conditions.
The overall objective is a weighted sum of these losses:
\begin{equation}
\mathcal{L}(\theta)= {\lambda}_{r}\mathcal{L}_{r}(\theta) + {\lambda}_{b}\mathcal{L}_{b}(\theta),
\end{equation}
where ${\lambda}_{r}$ and ${\lambda}_{b}$ are the non-negative weights of each loss term.
This training process naturally aligns with multi-task learning (MTL), as it seeks to minimize both losses simultaneously.

\begin{figure}[t]
    \centering
\includegraphics[width=0.95\linewidth]{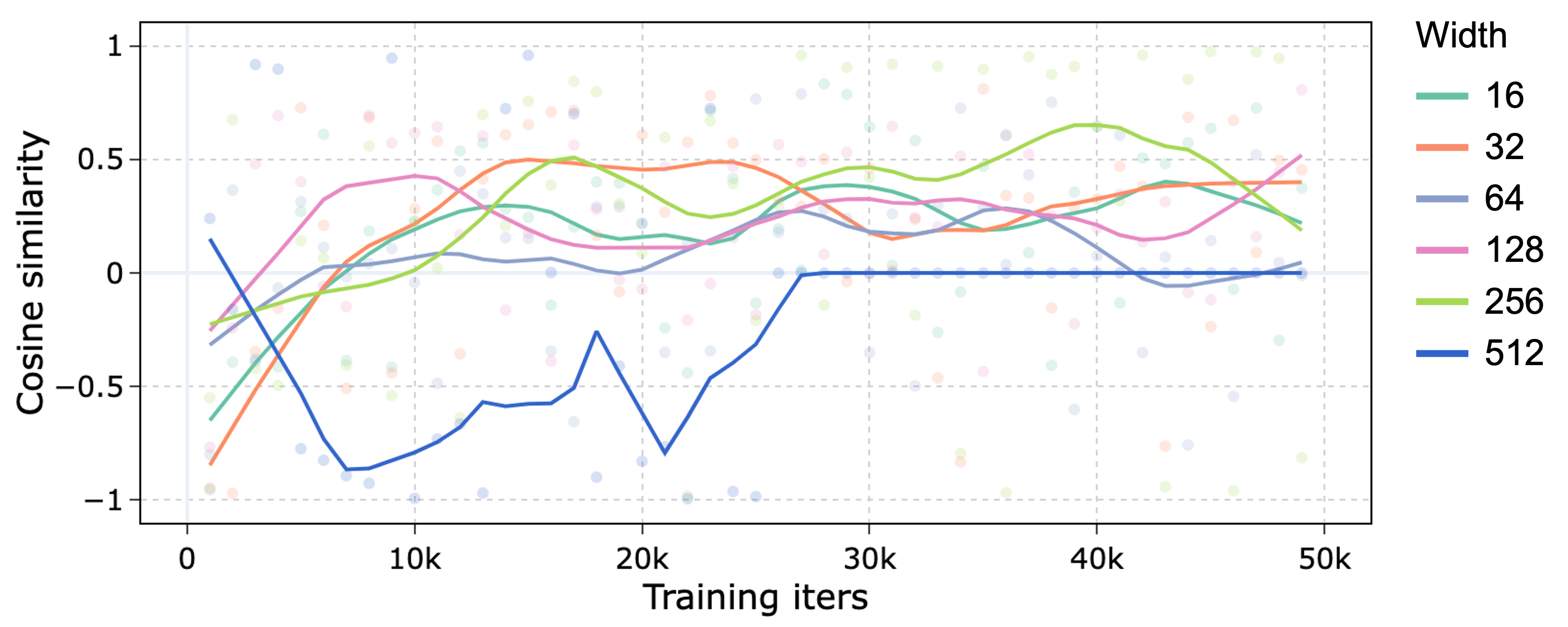}
\vspace{-0.1cm}
\caption{Cosine similarity of gradients versus training step for different network widths. Wide models converge near 0, indicating orthogonality of gradients and reduced cross-objective interaction.}
\vspace{-0.2cm}
\end{figure}

\noindent\textbf{Conflict-averse training} \cite{pcgrad, cagrad}, 
a key technique in MTL, addresses gradient conflicts between tasks.
These conflicts arise when the inner product of gradients is negative, \textit{i.e.}, $g_{i}^{T} g_{j} < 0, \forall i \neq j$, where $g_{i} = \nabla_{\theta} \mathcal{L}_{i}(\theta)$ represents the gradients of the $i$-th objective.
Such conflicts disrupt optimization by inducing destructive interference, ultimately degrading the performance of trained model.
The conflict-averse training mitigates this issue by adjusting gradients to align in cooperative directions, ensuring that tasks contribute constructively to the optimization process.

In the context of PINNs, gradient conflicts frequently occur between the residual loss $\mathcal{L}_{r}(\theta)$ and boundary loss $\mathcal{L}_{b}(\theta)$.
The most relevant works to ours are \cite{dcgd, config}, which adopt gradient manipulation strategies to alleviate these conflicts in PINNs.
Following them, we define the generalized gradient manipulation strategy as:
\begin{equation}
\hat{g}=\mathcal{A}\left(\nabla_{\theta} \mathcal{L}_r\left(\theta\right), \nabla_{\theta} \mathcal{L}_b\left(\theta\right)\right),
\end{equation}
where $\mathcal{A}$ represents an adjustment function tailored to reduce conflicts.
For example, in \cite{dcgd}, the adjustment function $\mathcal{A}$ can be defined as:
\begin{equation}
\begin{aligned}
& \hat{g}=\mathcal{A}(g_{r}, g_{b})=\frac{\left\langle g^{c}, \nabla_{\theta} \mathcal{L}\left(\theta\right)\right\rangle}{\left\|g^{c}\right\|^2} g^{c}, \quad \\
& \textit{ where } g^c=\frac{\nabla_{\theta} \mathcal{L}_r\left(\theta\right)}{\left\|\nabla_{\theta} \mathcal{L}_r\left(\theta\right)\right\|}+\frac{\nabla_{\theta} \mathcal{L}_b\left(\theta\right)}{\left\|\nabla_{\theta} \mathcal{L}_b\left(\theta\right)\right\|}.
\end{aligned}
\end{equation}
This adjustment function operates by projecting the overall gradient $\nabla_{\theta} \mathcal{L}(\theta)$ onto a direction $g^c$, which combines the normalized gradients of $\mathcal{L}_{r}$ and $\mathcal{L}_{b}$.
By aligning gradients, this manipulation effectively reduces conflicts, enabling more stable optimization convergence.
The adjusted gradient $\hat{g}$ is then used for parameter updates in gradient descent as $\theta_{t+1}=\theta_{t} - \eta \hat{g}$, where $\eta$ is the learning rate.

\begin{figure}[t]
\begin{center}
\begin{tabular}{c@{\;}c}
\includegraphics[width=0.475\columnwidth]{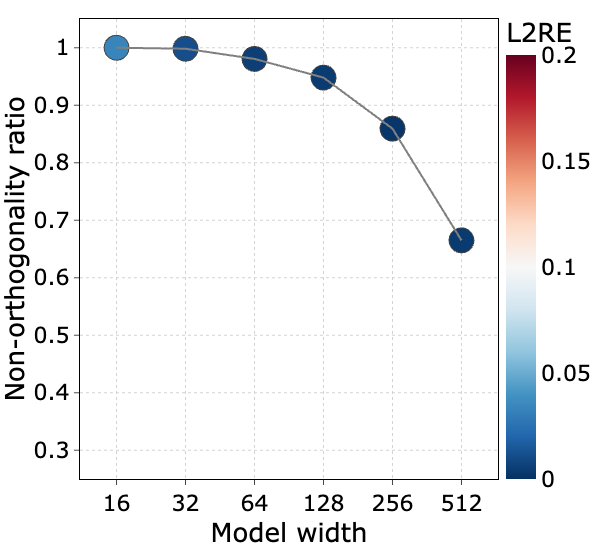} &
\includegraphics[width=0.475\columnwidth]{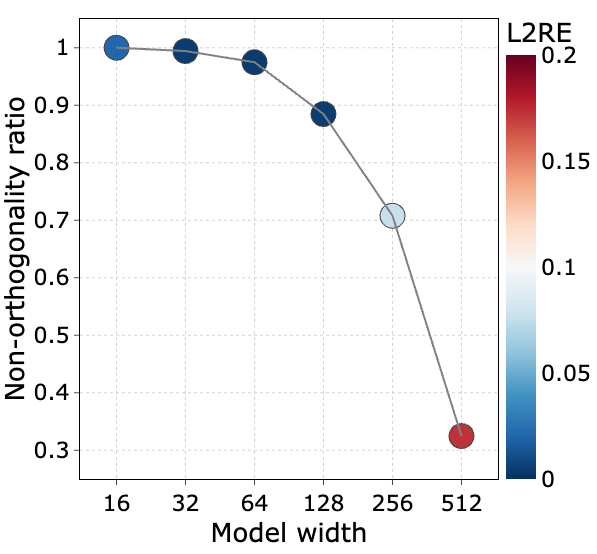} \\
(a) Vanilla PINNs  & (b) Conflict-averse PINNs \\
\end{tabular}
\end{center}
\vspace{-0.2cm}
\caption{Sample fraction with absolute dot product exceeding $\varepsilon$ ($\varepsilon=10^{-12}$). The ratio decreases with capacity and is further reduced under conflict-averse training. Color encodes relative L2 error.}
\vspace{-0.2cm}
\end{figure}
\subsection{Empirical Observations}
To investigate the side-effect of conflict-averse training, we conduct experiments on the viscous Burgers' equation, a widely used benchmark for evaluating PINN performance.
Using a plain multi-layer perceptron (MLP) trained with the Adam optimizer as a baseline, we compare the performance of conflict-averse method, adopting DCGD \cite{dcgd} as a representative benchmark.

Figure 2 reports the cosine similarity between the gradients of the residual and boundary losses, $\nabla_{\theta}\mathcal{L}_{r}$ and $\nabla_{\theta}\mathcal{L}_{b}$, as the network width increases while fixing depth to five layers.
As training progresses, the magnitude of negative alignment (conflict) generally diminishes.
Notably, for large width (512), the cosine similarity converges near zero, indicating an emergent orthogonality between the two gradients.
To quantify this effect, Figure 3 reports, over the training set, the fraction of samples for which $|\nabla_{\theta}\mathcal{L}_{r} \cdot \nabla_{\theta}\mathcal{L}_{b}| > \varepsilon$ with $\varepsilon=10^{-12}$.
This ratio decreases as model capacity grows, and the decrease is amplified under conflict-averse training, implying that sample-wise gradients increasingly become orthogonal.

These observations suggest that larger models adopt a capacity-enabled shortcut: gradients decouple toward orthogonality, consistent with the emergence of objective-exclusive functional modularity.
Moreover, conflict-averse training accelerates this decoupling, reinforcing modular segregation rather than fostering cross-objective interaction.
\begin{table}[t]
\centering
\begin{tabular}{p{0.97\columnwidth}}
\toprule
\textbf{Algorithm 1} Optimization procedure of the proposed framework \\ \midrule
\textbf{Input:} PINN model $u(\cdot;\theta, t_{r}, t_{b})$; residual/boundary sets $\Omega$, $\partial\Omega$; \\
\hspace*{\algorithmicindent}\quad\, sharpness factor $\beta$; learning rates $\eta_{\theta}$, $\eta_{t}$.
\begin{algorithmic}[1]
\State \textbf{Initialize} a set of model parameters $\theta$ (Xavier initialization), \\
\hspace*{\algorithmicindent}\qquad\, and threshold vectors $t_r \leftarrow \boldsymbol{0}$ and $t_b \leftarrow \boldsymbol{0}$.
\While{\textit{model not converge}}

    \For{$iter=1$ to $I_{\text{epoch}}$}

        \State {\textbf{Sample}} batches of $N_{r} \subset \Omega$ and $N_{b} \subset \partial\Omega$
        \State $M \leftarrow S(|\theta|-t)$, where $t=(t_{r} + t_{b}) / 2$
        \State $\tilde{\theta} \leftarrow M \odot \theta$
        \State {\textbf{Compute}} losses on masked weights $\mathcal{L}_r(\tilde{\theta})$, $\mathcal{L}_b(\tilde{\theta})$
        \State $\hat{g} \leftarrow \mathcal{A}(\nabla_{\theta} \mathcal{L}_r(\tilde{\theta}), \nabla_{\theta} \mathcal{L}_b(\tilde{\theta}))$
        \State \textbf{Update} $\theta \leftarrow \theta - \eta_{\theta} \cdot \hat{g}$
        \State {\textbf{Compute}} the loss on threshold vectors $\mathcal{L}_{s}(t_{r}, t_{b})$
        \State $g_{t_r} \leftarrow \nabla_{t_r}(\mathcal{L}_{r}(\tilde{\theta})+\mathcal{L}_{s}(t_{r}, t_{b}))$
        \State $g_{t_b} \leftarrow \nabla_{t_b}(\mathcal{L}_{b}(\tilde{\theta})+\mathcal{L}_{s}(t_{r}, t_{b}))$
        \State \textbf{Update} $t_{r} \leftarrow t_{r} - \eta_{t} \cdot g_{t_r}$
        \State \textbf{Update} $t_{b} \leftarrow t_{b} - \eta_{t} \cdot g_{t_b}$

    \EndFor
    
\EndWhile

\end{algorithmic}
\textbf{Output:} trained model weights $\theta$ and threshold vectors $t_{r}$, $t_{b}$ \\
\bottomrule
\end{tabular}
\end{table}
\section{Our Approach}
As noted earlier, conflict-averse training in high-capacity PINNs can induce a capacity-enabled shortcut—objective-exclusive functional modularity—where the residual and boundary objectives decouple and gradients drift toward orthogonality.
To counter this effect, we propose Modular-Sparsity Synchronization (ModSync), a model-agnostic framework that augments conflict-averse training with structural optimization.
ModSync has two key components: (i) identification of objective-specific modules, and (ii) regularization of objective-exclusive connections.
In the following sections, we describe the specific mechanisms of the two aspects, and then present the final optimization procedure that combines these strategies.

\subsection{Identifying Objective-Specific Modules}
We treat the residual and boundary objectives as distinct tasks and aim to identify subnetworks that each objective predominantly develops on its own.
To this end, we adopt ideas from dynamic sparse training (DST) \cite{dst, bidst, rigl, dst_2} to learn binary connectivity masks jointly with the network weights.

Let the parameters of the network consist of learnable weights $\theta$ and a binary mask $M$ that selects active connections.
For each layer $i$ with the weight matrix $\theta^{i} \in \mathbb{R}^{d_{o} \times d_{i}}$, we associate a trainable threshold vector $t^{i} \in \mathbb{R}^{d_{o}}$ (one threshold per output unit).
Using a step function $S(\cdot)$, the binary mask is constructed as:
\begin{equation}
\begin{aligned}
& q^{i j}=|\theta^{i j}|-t^{i}, \\
& M^{i j}=S(q^{i j}), \textit{ where } 1 \leq i \leq d_o, 1 \leq j \leq d_i.
\end{aligned}
\end{equation}
With the dynamic mask $M$, an element $M^{i,j}$ is set to zero if $\theta^{i,j}$ is pruned, resulting in the sparse parameters $M \odot \theta$, where $\odot$ is the Hadamard product.
The sparse parameters are updated through gradient-based optimization.
Since $S(\cdot)$ is non-differentiable, we approximate it with a smooth gradient estimator, leveraging the long-tailed estimator \cite{dst}.

To reveal objective-specific structure while keeping a single forward connectivity, we decompose the thresholds as $t=(t_{r} + t_{b})/2$, where $t_{r}$ and $t_{b}$ are updated using gradients from the residual and boundary objectives, respectively.
The forward propagation uses the shared mask built from the composite vector $t$, ensuring a common inference pathway, whereas the backward propagation applies separate updates to $t_{r}$ and $t_{b}$ with their own objective gradients (the weights continue to follow the conflict-averse update).

Combined with the sparsity regularizer introduced next, the threshold vectors $t_{r}$ and $t_{b}$ are optimized to suppress connections that are redundant for their respective objective.
We therefore regard each threshold vector as an implicit objective-specific sparse subnetwork.
Based on these, we quantify objective exclusivity by the dissimilarity between their supports, which will serve as the target signal for the sparsity-synchronization regularizer in the next subsection.

\subsection{Regularizing Modular-Sparsity}

Building on the objective-specific modules, we introduce a sparsity regularizer to suppress unexpected objective exclusivity while retaining interaction-promoting connections.
The regularizer is designed to satisfy two requirements: \textit{i)} encourage each objective to prune redundant connections (promote sparsity), and \textit{ii)} de-emphasize the sparsity penalty on connections that are likely shared across objectives so as to preserve interaction.
Specifically, the proposed sparsity term is defined as:
\begin{equation}
\mathcal{L}_s(t_r, t_b)=\sum_{i=1}^C \mathcal{R}_w(t_r^i, t_b^i) \cdot (\mathcal{R}_s(t_r^i)+\mathcal{R}_s(t_b^i)),
\end{equation}
where $C$ represents the total number of layers in the training model.

To fulfill the first requirement, we adopt a monotonically decreasing sparsity-inducing function $\mathcal{R}_{s}(t)=\exp(-t)$, inspired by original DST \cite{dst}.
It penalizes small $t$ to prune redundant connections while avoiding over-sparsification pressure at large $t$, yielding binary masks that focus capacity on objective-relevant structure.
Formally, the sparsity-inducing function is given as:
\begin{equation}
\mathcal{R}_s(t_r^i)=\exp (-t_r^i), \quad \mathcal{R}_s(t_b^i)=\exp (-t_b^i).
\end{equation}

To satisfy the second requirement, we introduce a reweighting function $\mathcal{R}_w$, which dynamically adjusts the sparsity penalty based on the L2 distance between $t_r^i$ and $t_b^i$.
This function assigns higher weights to elements with significant differences, emphasizing task-specific connections, while suppressing sparsity penalties for shared connections:
\begin{equation}
\mathcal{R}_w(t_r^i, t_b^i)=\exp (|t_r^i-t_b^i|^2 /~ \beta),
\end{equation}
where the sharpness factor $\beta \in [0, 1]$ controls the sensitivity of the weighting function.
To ensure training stability, we define this function in exponential form, while setting the scaling factor $\lambda_s$ in Eq. 10 to a relatively small value.

\subsection{Proposed Overview}
Consequently, the overall objective function of the proposed ModSync can be defined as:
\begin{equation}
\mathcal{L}(\theta,t_r,t_b)= {\lambda}_{r}\mathcal{L}_{r}(\tilde{\theta}) + {\lambda}_{b}\mathcal{L}_{b}(\tilde{\theta})+{\lambda}_{s}\mathcal{L}_{s}(t_r,t_b),
\end{equation}
where $\tilde{\theta}=M \odot \theta$ represents a sparse parameters with a binary mask applied.
For optimization with respect to $\theta$, we use the refined gradient obtained by applying the adjustment function defined in Eq. 4 to the gradients derived from $\mathcal{L}_{r}$ and $\mathcal{L}_{b}$.
In contrast, $t_{r}$ and $t_{b}$ re optimized by applying independent gradients for the two objectives.
The overall training procedure of the proposed method can be found in Algorithm 1.
\begin{table*}[t]
\begin{center}
\caption{Comparison of PINN performance across multiple PDE benchmarks.}
\vspace{0.1cm}
\resizebox{0.85\linewidth}{!}
{
\begin{tabular}{lcccccccccc}
\toprule
 & \multicolumn{2}{c}{Helmholtz (2D)} & \multicolumn{2}{c}{Helmholtz (3D)} & \multicolumn{2}{c}{Klein-Gordon (2D)} & \multicolumn{2}{c}{Klein-Gordon (3D)} & \multicolumn{2}{c}{Burgers} \\
Methods & Absolute & Relative & Absolute & Relative & Absolute & Relative & Absolute & Relative & Absolute & Relative \\
\midrule
Adam      & 3.701E-01 & 0.748\% & 2.380E+01 & 38.99\% & 5.146E-01 & 1.200\% & 5.298E+00 & 4.496\% & 9.926E-02 & 0.101\% \\
LRA       & 5.723E-01 & 1.156\% & 5.737E+01 & 94.02\% & 5.460E-01 & 1.274\% & 7.904E+00 & 6.707\% & 6.251E-01 & 0.636\% \\
NTK       & 6.518E-01 & 1.317\% & 2.616E+01 & 43.87\% & 7.108E-01 & 1.658\% & \textbf{3.177E+00} & \textbf{2.695\%} & 4.628E-01 & 0.471\% \\
PCGrad    & 3.311E+01 & 66.89\% & 3.729E+01 & 61.11\% & 3.743E-01 & 0.873\% & 4.526E+00 & 3.841\% & 2.913E-01 & 0.296\% \\
CAGrad    & 3.465E-01 & 0.700\% & 2.290E+01 & 37.53\% & 2.337E-01 & 0.545\% & 5.294E+00 & 4.492\% & 2.867E-01 & 0.291\% \\
MultiAdam & 5.108E+00 & 10.32\% & 3.811E+01 & 62.45\% & 1.882E+01 & 43.91\% & 2.729E+01 & 23.15\% & 1.313E+01 & 13.35\% \\
\midrule\midrule
DCGD      & 1.240E+01 & 25.05\% & 5.190E+01 & 85.05\% & 9.525E-02 & 0.222\% & 4.780E+00 & 4.056\% & 3.357E+01 & 34.15\% \\
w./ Ours  & \underline{4.774E-02} & \underline{0.096\%} & \underline{2.002E+01} & \underline{32.81\%} & \underline{8.194E-02} & \underline{0.191\%} & \underline{4.274E+00} & \underline{3.627\%} & \textbf{3.208E-02} & \textbf{0.033\%} \\
\midrule
ConFIG    & 1.242E+01 & 25.09\% & 5.581E+01 & 91.45\% & 1.112E-01 & 0.259\% & 6.784E+00 & 5.756\% & 6.075E+01 & 61.80\% \\
w./ Ours  & \textbf{3.836E-02} & \textbf{0.077\%} & \textbf{1.986E+01} & \textbf{32.54\%} & \textbf{6.996E-02} & \textbf{0.163\%} & 5.501E+00 & 4.668\% & \underline{4.150E-02} & \underline{0.042\%} \\
\bottomrule
\end{tabular}
}
\end{center}
\vspace{-0.8cm}
\end{table*}
\begin{figure}[t]
\begin{center}
\begin{tabular}{c@{\quad}c}
\includegraphics[width=0.45\columnwidth]{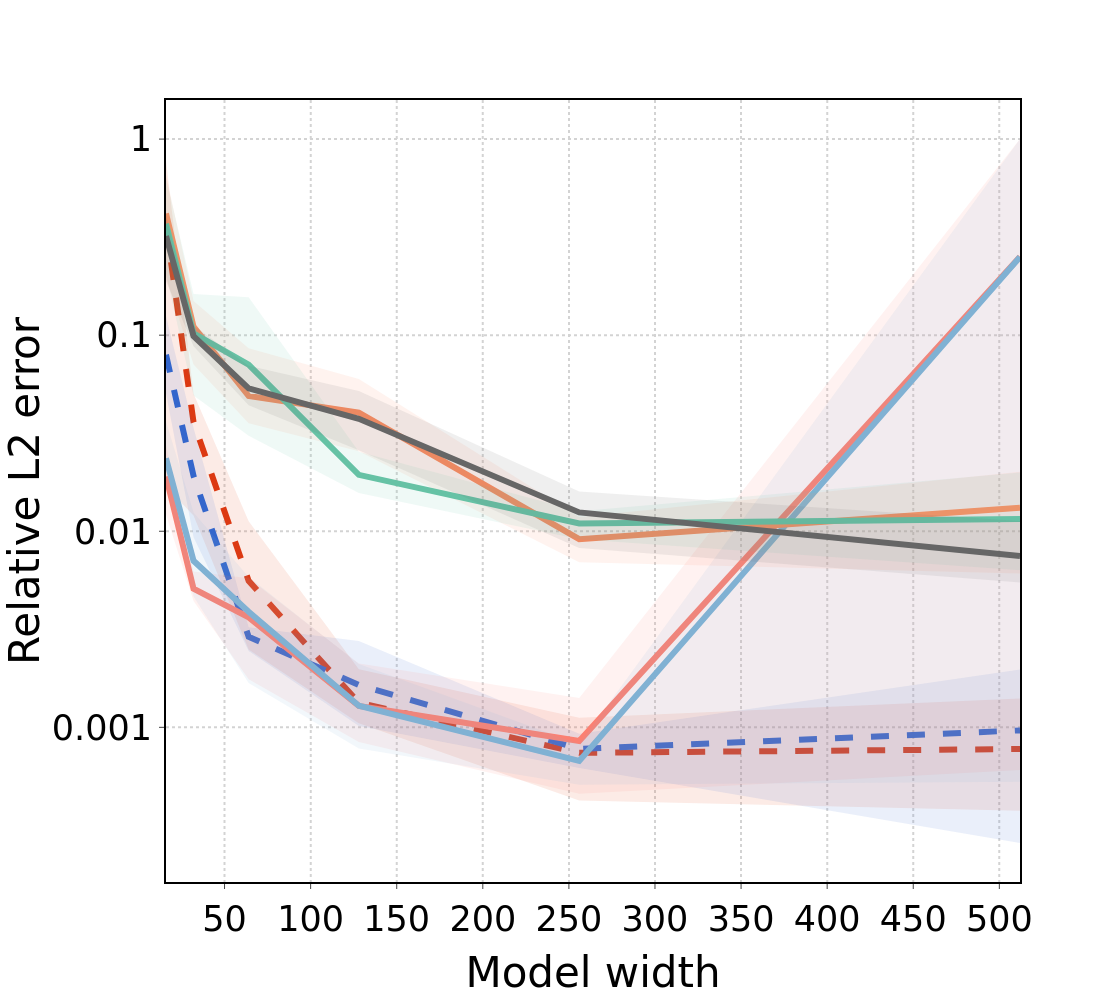} &
\includegraphics[width=0.45\columnwidth]{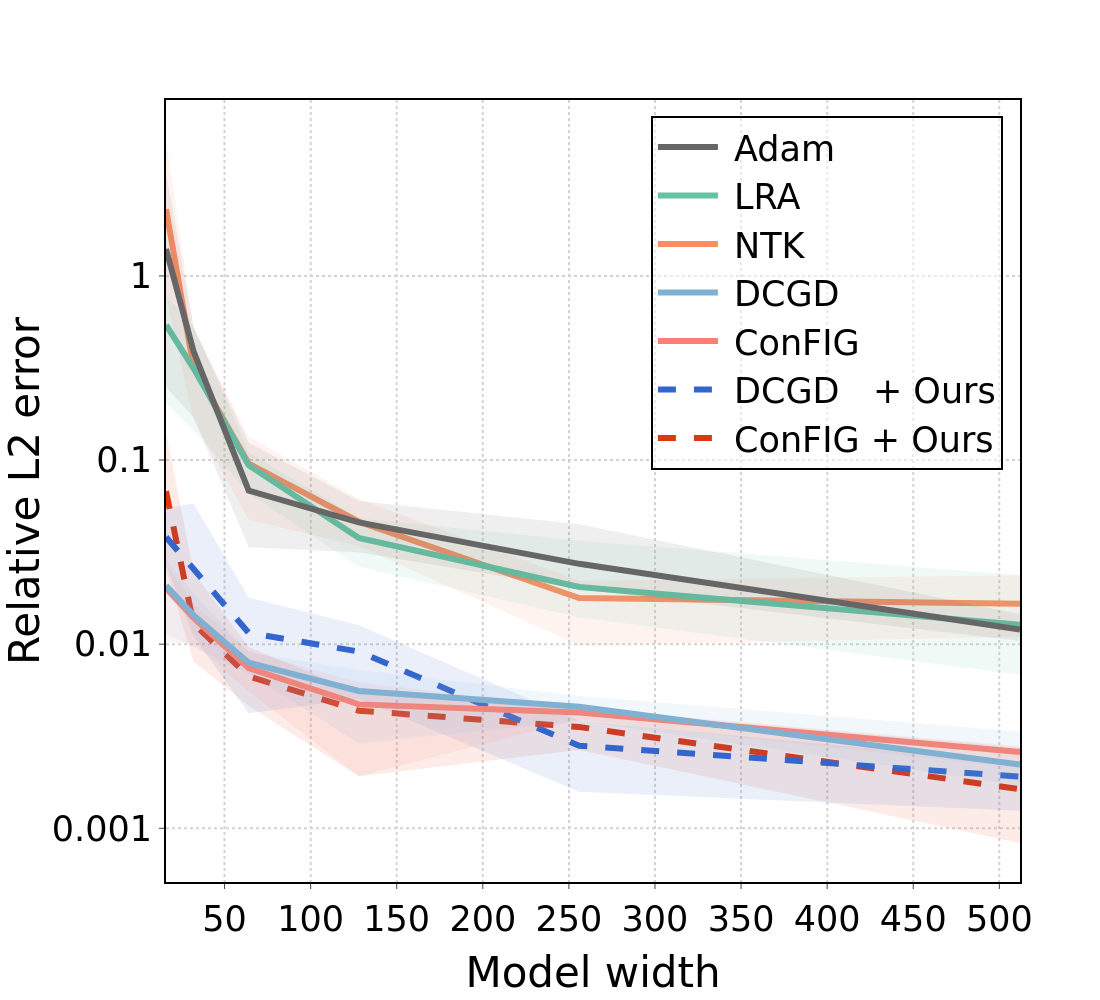} \\
(a) Helmholtz (2D)  & (b) Klein-Gordon (2D) \\
\end{tabular}
\end{center}
\vspace{-0.3cm}
\caption{Effect of model capacity on PINN performance across diverse PDE benchmarks. The proposed ModSync consistently prevents capacity-induced failure modes and achieves lower relative errors than existing baselines.}
\vspace{-0.3cm}
\end{figure}
\section{Experiments}
We evaluate the proposed method on several challenging PDEs to assess its training stability and accuracy. 
Unless otherwise noted, models train for 50{,}000 iterations; results are the average of three different seeds, selecting the best validation checkpoint.
We use an identical set of hyperparameters for all ModSync runs: $\{\lambda_r=1.0, \lambda_b=1.0, \lambda_s=0.0001, \eta_{\theta}=0.001, \eta_{t}=0.0001, \beta=0.3\}$.
Detailed PDE formulations and sampling strategies follow \cite{dcgd, config, multiadam}.

\subsection{Comparison on Benchmark Equations}
We consider three popular 2D benchmarks—Helmholtz, Klein–Gordon, and Burgers’—and additionally evaluate Helmholtz and Klein–Gordon in 3D. Baselines include Vanilla PINNs (Adam) and PINN-specific optimization methods (LRA \cite{lra}, NTK \cite{ntk}, MultiAdam \cite{multiadam}), as well as MTL gradient-manipulation methods (PCGrad \cite{pcgrad}, CAGrad \cite{cagrad}). To test compatibility, ModSync is integrated into two conflict-averse PINN approaches, DCGD \cite{dcgd} and ConFIG \cite{config}.
Table 1 reports absolute and relative errors for a 5-layer MLP with width 512. We observe that several conflict-averse methods (e.g., PCGrad, DCGD, ConFIG) can enter capacity-induced failure modes. In contrast, integrating ModSync consistently mitigates these failures and achieves state-of-the-art accuracy on most benchmarks.

\subsection{Verification of Scalability Effects}
To directly probe ModSync’s suppression of functional modularity, we study model width scaling. Guided by Table 1, we adopt strong baselines—Adam, LRA, NTK, DCGD, and ConFIG—and evaluate ModSync when combined with DCGD and ConFIG. All models use a 5-layer MLP; only width varies.
Figures 1 and 4 show that DCGD and ConFIG perform well at small widths but degrade as capacity increases, exhibiting the predicted failure modes. When augmented with ModSync, both methods retain stable convergence and competitive accuracy even at larger widths. These results substantiate that ModSync effectively counters the structural shortcut (functional modularity) responsible for capacity-driven failures.

\subsection{Ablation Study}
To assess the contribution of each component in ModSync, we conduct ablations on the Burgers’ equation using a 5-layer PINN (width 512). Table 2 summarizes the results.
Compared to standard PINNs (Adam), the DCGD baseline fails to converge, yielding a large relative error.
Introducing a plain DST variant with only the sparsity regularizer $\mathcal{R}_s$ (third row) restores convergence, and adopting objective-centric DST \cite{dst} with dual thresholds (fourth row) further improves accuracy.
We attribute these gains to early-stage correction of redundant connections that otherwise foster functional modularity.
Finally, the full ModSync—combining dual thresholds, $\mathcal{R}_s$, and the reweighting $\mathcal{R}_w$ —achieves the best performance, delivering meaningful improvements over both Adam-optimized PINNs and DCGD-optimized models.

\begin{table}[t]
\begin{center}
\caption{Ablation studies of the individual components.}
\resizebox{0.95\columnwidth}{!}
{
\begin{tabular}{ccccc|cc}
\toprule
DCGD & single $t$ & dual $t$ & $\mathcal{R}_s$ & $\mathcal{R}_w$ & Relative (\%) & GPU (hours) \\ \midrule
            &             &             &             &             & 0.101\% & 0.202 \\ 
\checkmark  &             &             &             &             & 34.15\% & 0.274 \\ \midrule
\checkmark  & \checkmark  &             & \checkmark  &             & 0.304\% & 0498 \\
\checkmark  &             & \checkmark  & \checkmark  &             & 0.286\% & 0.561 \\
            &             & \checkmark  & \checkmark  & \checkmark  & 0.093\% & 0.450 \\
\checkmark  &             & \checkmark  & \checkmark  & \checkmark  & \textbf{0.033\%} & 0.587 \\
\bottomrule
\end{tabular}
}
\end{center}
\vspace{-0.5cm}
\end{table}
\section{Conclusion}
%
%
We investigated the fragility of conflict-averse training in PINNs and showed that growing model capacity induces functional modularity, where networks partition into task-exclusive modules and suppress cross-objective interaction.
To address this, we proposed ModSync, a structural optimization that integrates with conflict-averse schemes by penalizing exclusive connections while preserving interaction pathways.
Experiments across diverse PDEs demonstrated that ModSync mitigates capacity-induced failures, sustains stable convergence, and achieves state-of-the-art accuracy.
These results underscore the role of structural regularization in enabling scalable, reliable PINN training.
%
\newpage
\bibliographystyle{IEEEbib}
\bibliography{icassp2026}
\end{document}